\pdfoutput=1
\documentclass[letterpaper, 10 pt, conference]{ieeeconf}  

\IEEEoverridecommandlockouts                              

\usepackage{times}
\usepackage{epsfig}
\usepackage{graphicx}
\usepackage{amsmath}
\usepackage{amssymb}
\usepackage{bbm}

\usepackage{cite}
\usepackage{float}
\usepackage{algorithm}
\usepackage{algorithmicx}
\usepackage{algpseudocode}
\usepackage{mathtools}
\usepackage{url}

\usepackage[table]{xcolor} 
\usepackage[para,online,flushleft]{threeparttable}
\usepackage{booktabs}
\usepackage{longtable}
\usepackage{tablefootnote}
\usepackage{etoolbox}
\appto\TPTnoteSettings{\footnotesize}

\usepackage{arydshln}
\usepackage{subcaption}
\usepackage{makecell}
\usepackage{pifont}
\usepackage{multirow}
\usepackage{booktabs}
\usepackage{chngcntr}
\usepackage{float}
\usepackage{diagbox}

\usepackage{amsmath}
\usepackage{romannum}

\makeatletter
\newcommand\threepart@subtable{
  \caption@setoptions{threepartsubtable}%
  \caption@ORI@threeparttable
}

\makeatother

\makeatletter
\let\NAT@parse\undefined
\makeatother
\usepackage[breaklinks=true,bookmarks=false]{hyperref}
\hypersetup{
    colorlinks=true,
    linkcolor=blue,
    filecolor=magenta,
    urlcolor=cyan,
}

\title{\LARGE \bf
Autonomous Vehicles that Alert Humans to Take-Over Controls: Modeling with Real-World Data
}

\author{Akshay Rangesh$^{\dag, 1}$, Nachiket Deo$^{\dag, 1}$, Ross Greer$^{\dag, 1}$, Pujitha Gunaratne$^2$ and Mohan M. Trivedi$^1$\\
{\tt\small \{arangesh, ndeo, regreer, mtrivedi\}@ucsd.edu, pujitha.gunaratne@toyota.com}
\thanks{
\hspace{-0.27cm}$^\dag$authors contributed equally
\newline$^1$Laboratory for Intelligent \& Safe Automobiles, UC San Diego
\newline$^2$Toyota CSRC
}
}

\begin{document}

\maketitle
\thispagestyle{empty}
\pagestyle{empty}

\begin{abstract}
With increasing automation in passenger vehicles, the study of safe and smooth occupant-vehicle interaction and control transitions is key. In this study, we focus on the development of contextual, semantically meaningful representations of the driver state, which can then be used to determine the appropriate timing and conditions for transfer of control between driver and vehicle. To this end, we conduct a large-scale real-world controlled data study where participants are instructed to take-over control from an autonomous agent under different driving conditions while engaged in a variety of distracting activities. These take-over events are captured using multiple driver-facing cameras, which when labelled result in a dataset of control transitions and their corresponding take-over times (TOTs). We then develop and train TOT models that operate sequentially on mid to high-level features produced by computer vision algorithms operating on different driver-facing camera views. The proposed TOT model produces continuous predictions of take-over times without delay, and shows promising qualitative and quantitative results in complex real-world scenarios.
\end{abstract}


\section{Introduction}

\begin{figure}[t]
    \center{\includegraphics[width=0.9\linewidth]
    {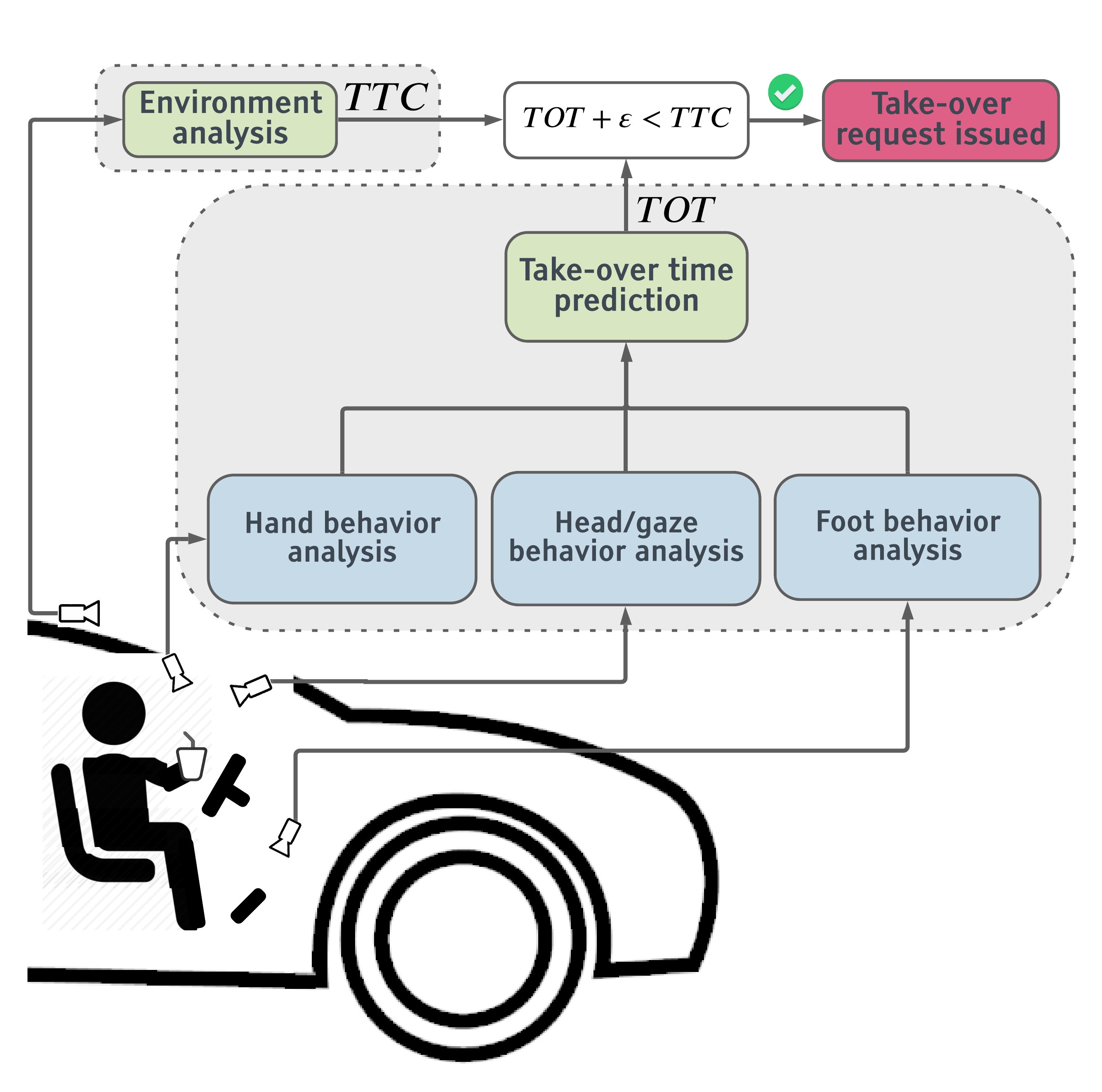}}
    \caption{Illustration of the proposed take-over time (TOT) prediction approach using driver-facing cameras. The governing logic in an autonomous vehicle must understand and model driver behavior continuously in order to ensure safe and smooth transfer of control between automation and humans.}
  \label{fig:motivation}
\end{figure}

Control transitions in the context of partially automated vehicles entail the transfer of vehicle controls from the autonomous agent to the human driver and vice versa. In this study, we primarily focus on transitions from the autonomous agent to the human driver - as these scenarios are inherently risky and require timely human intervention to avoid collision. In describing such control transitions, we make use of the take-over time (TOT) metric, defined as the interval of time between a take-over request (TOR) being issued and the assuming of human control. The take-over request could be an auditory/visual/tactile cue used to indicate to the driver that their intervention is immediately needed. Due to the complexity of human attention, we define the assumption of control as the completion of the following three behaviors:
\begin{enumerate}
    \item Hands-on-wheel: hand(s) return to the vehicle's steering control.
    \item Foot-on-pedal: foot returns (from floorboard or hovering) to make contact with any driving pedal.
    \item Eyes-on-road: gaze is directed forward, toward the active driving scene. 
\end{enumerate}
We make the assumption that these three cues occurring simultaneously are sufficient to consider the driver both attentive to the scene and in control of the vehicle. 

As depicted in Fig.~\ref{fig:motivation}, the transition of control from an autonomous agent to the human driver should be a function of both the surrounding scene and the state of the driver. The surrounding scene can be concisely expressed using a metric such as time-to-collision (TTC), whereas the state of the driver can be captured by the predicted TOT. Combined, this forms a criterion for safe control transitions: 
\begin{equation}
    TOT + \varepsilon < TTC,
\end{equation}
where $\varepsilon$ is a marginal allowance for human maneuvering. A system that takes the state of the driver into account can decide between handing over control if the driver is ready, versus coming to a safe and smooth halt if not. 

While there are many approaches to predict TTC, TOT prediction (especially in the real world) remains relatively unexplored. This study is the first to train TOT prediction models using in-cabin sensors that monitor human subjects in real (non-simulated) driving scenarios. By using real-world data, we train and develop a sequential neural network based on LSTMs to accurately predict TOTs for drivers engaged in a variety of secondary tasks. In conjunction with a TTC model, the proposed TOT prediction model could allow the governing logic to make informed decisions during control transitions.

\section{Observing the Driver for Take-over Time Prediction}

As stated in \cite{daily2017self}, ``...the design of intelligent driver-assistance systems, especially those that activate controls of the car to prevent accidents, requires an accurate understanding of human behavior as well as modeling of human-vehicle interactions, driver activities while behind the wheel, and predictable human intent." Without understanding and cooperating with the driver state, take-over requests may be issued with inadequate time to execute manual take-over. A comprehensive, cross-disciplinary survey of driver attention modeling, \cite{kotseruba2021behavioral} introduces a progression of approaches toward understanding in-cabin factors which are immediately relevant in the context of control transitions from automated to manual control.  

In most previous studies, control transitions in autonomous vehicles have been framed around particular moments expected during take-over events. A transition begins with a take-over request, to which the driver will react. At some point, the driver will have established control, traditionally evidenced by (and also evaluated by) the quality of performance metrics such as braking patterns, speed, and steering wheel angle. 

Many studies in the realm of human factors and human-robot interaction have sought to understand and improve autonomous vehicle control transitions, evaluating the effects of take-over request methodologies, sample demographics, and secondary non-driving-related tasks. Due to possible catastrophic failure cases, video simulation is used in lieu of real-world driving for many of these studies. In \cite{eriksson2017takeover}, researchers analyzed the take-over time of participants in a Highly Automated Driving simulator. Participants may have been assigned a secondary task, such as reading a magazine, before being presented with an audio-visual take-over request, shown to correspond to significantly longer control transition times. \cite{dogan2017transition} found that when take-over requests were correlated with traffic scene cues such as vehicle speed and traffic density, the anticipatory scene information led to quicker participant reactions to take-over requests. They make a distinction between driver reaction and driver control, the latter (measured as visual attention and lane deviation) showing no significant change whether take-over requests were scene-informed or sporadic. Looking inside the cabin, studies such as \cite{du2020evaluating, ma2020promote, kaye2021young, clark2017age} explore the effects of demographics, cognitive load of secondary tasks, and nature of take-over request on driver response time and quality. While in-cabin activity has been classified using machine learning approaches, as in \cite{roitberg2020cnn, rangesh2018exploring, ohn2015surveillance, cheng2007multi}, there is additional importance in using this information to assess driver readiness and predict take-over times. 

Most indicators studied in previous works are non-predictive by nature, as these take-over time or quality signals appear following the instance of take-over. Rather than measuring human reaction time in simulated data, in our work, we novelly explore the gray area that exists between initial reaction and control by analyzing human behavior in the frames between the issued take-over and brake or steering action. Contrasting previous methods, we show that by using non-intrusive visual modalities, it is possible to observe a foot at-the-ready to brake without measuring 10\% pedal depression, or ready hands-on-the-wheel without a 2-degree steering jitter. This ensemble of visual driver readiness features, developed in \cite{tawari2014driver} has been shown to closely match human expectation of vehicle control in \cite{deo2019looking}, so it is a natural extension to use these cues for the purpose of take-over time prediction. 

Further, our work is unique in its use of real-world driving data, as opposed to the more widely-studied video simulations. While studies such as \cite{naujoks2019noncritical} have replaced video simulation with Wizard-of-Oz driving and others such as \cite{martin2019drive} have explored secondary activities while driving or supervising, our work is first to feature a vehicle in exclusive, live control of the driver and AV driver-assistance system during control transitions. Additionally, our data is representative of a wide range of secondary tasks, which vary in both duration and cognitive load. By combining the visual readiness features extracted from naturalistic driving data, we show an effective data-driven approache for predicting take-over time.

\section{Real-World Dataset \& Labels}

\subsection{Controlled Data Study (CDS)}
To capture a diverse set of real-world take-overs, we conduct a large-scale study under controlled conditions. More specifically, we enlist a representative population of 89 subjects to drive a Tesla Model S testbed mounted with three driver-facing cameras that capture the gaze, hand, and foot activity of the driver. In this controlled data study (CDS), we required each subject to drive the testbed for approximately an hour in a pre-determined section of the roadway, under controlled traffic conditions. During the drive, each test subject is asked to undertake a variety of distracting secondary activities while the autopilot is engaged, following which an auditory take-over request (TOR) is issued at random intervals. This initiates the control transition during which the driver is instructed to take control of the vehicle and resume the drive. Each such transition corresponds to one take-over event, and our CDS produces 1,375 take-over events in total. We further augment our dataset by randomly moving the TOR to a time after the actual TOR but before the take-over is initiated. The target times for these samples are adjusted accordingly.

\subsection{Labelling Procedure}
\noindent\underline{\textbf{Automated video segmentation:}} Each driving session is first segmented into 30 second windows surrounding take-over events, consisting of 20 seconds prior to the take-over request (TOR) and 10 seconds after the take-over event. The times at which these TORs were initiated are known and help us locate all take-over events from a driving session. 

\noindent\underline{\textbf{Event annotations:}} 
For each 30 second clip corresponding to a take-over event, we annotate the following:
\begin{enumerate}
    \item \textbf{Eyes-on-road:} We mark the first frame after the take-over request when the driver's eyes are on the road.
    \item  \textbf{Hands-on-wheel:} We mark the first frame after the take-over request when the driver's hands are on the wheel.
    \item \textbf{Foot-on-pedal:} We mark the first frame after the take-over event when the driver's foot is on the pedals.
    \item \textbf{Driver secondary activity:} We label the secondary activity being performed by the driver during each take-over event. We assign one of 8 possible activity labels: (1) No secondary activity (i.e. attentive), (2) talking to co-passenger, (3) looking at lap/eyes closed, (4) texting, (5) phone call, (6) using infotainment unit, (7) counting change, (8) reading a book or magazine.
\end{enumerate}

\subsection{Take-over Time Statistics}
\label{sec:d7.1-labels}

Of the 1,375 control transitions obtained from the CDS, 308 correspond to performing no secondary activity (attentive), 182 correspond to the driver talking to co-passengers, 85 correspond to drivers with their eyes closed or looking at their lap, 262 correspond to the driver texting, 42 correspond to the driver being on a phone call, 299 correspond to the driver interacting with the infotainment unit, 97 correspond to the driver counting coins, and finally 100 correspond to the driver reading a book or magazine. 

\begin{figure}[t]
\centering
\includegraphics[width=0.99\linewidth]{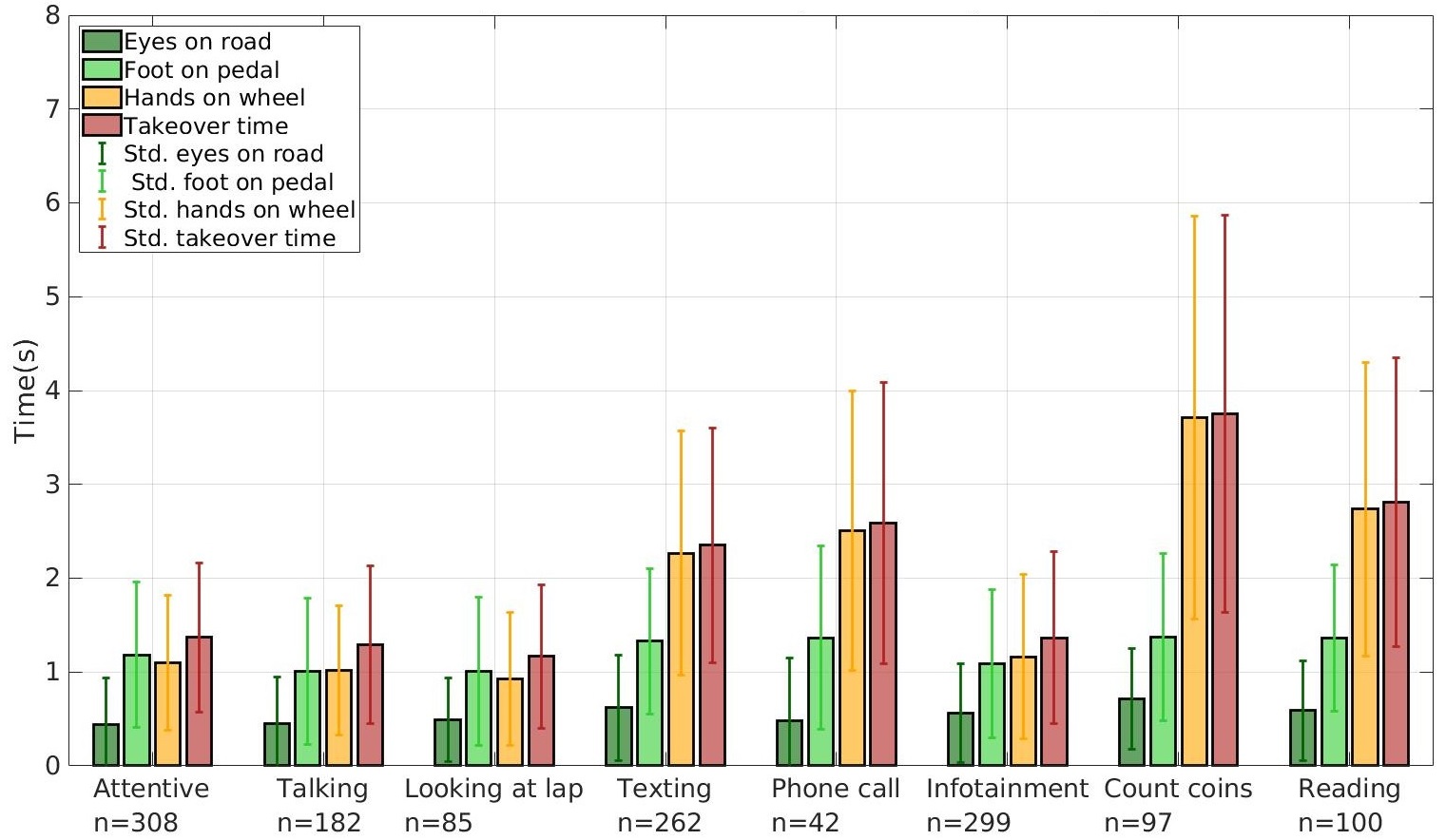}
\caption{Take-over time statistics from the CDS. We plot the mean values (with deviation bars) of the different take-over related event timings for each secondary activity.}
\label{fig:d7.1-totstats}
\end{figure}

Fig.~\ref{fig:d7.1-totstats} shows the average times corresponding to eyes-on-road, hands-on-wheel, and foot-on-pedal for each of the 8 secondary activities. It also shows the overall take-over time, defined as the maximum of the three markers for each event.
We note that texting, phone-calls, counting change and reading correspond to longer average take-over times, as compared to talking to the co-passenger or using the infotainment unit, which can be reasonably expected.
Counter to intuition, the `eyes closed/looking at lap' activity has low take-over times. This is mainly because the drivers are merely pretending to be asleep, since actual sleep could not have been achieved given the duration and nature of each trial. We also note that the 'hands-on-wheel' event seems to take much longer on average, as compared to eyes-on-road or foot-on-pedal. This reinforces the need for driver hand analysis. Finally, we note that for the more distracting secondary activities (reading, texting, phone calls, counting change), even the foot-on-pedal times are longer compared to the other secondary activities, although the secondary activities do not involve the driver's feet. Thus, there seems to be a delay corresponding to the driver shifting attention from secondary activity to the primary activity of driving.


\section{Take-over Time Prediction Model}

\subsection{Overview}\label{sec:d5-tot}
It is important to preserve both the diverse and sequential nature of all features related to driver behavior while designing a holistic take-over time (TOT) prediction framework. High level tasks such as TOT prediction are influenced by low level driver behaviors, both in the short and medium to long term. With this in mind, we propose the overall framework illustrated in Fig.~\ref{fig:d5-framework_tot} comprising of 4 unique blocks (or stages). We provide a brief description of each block below:

\begin{figure}[h]
  \centering
  \includegraphics[width=\linewidth]{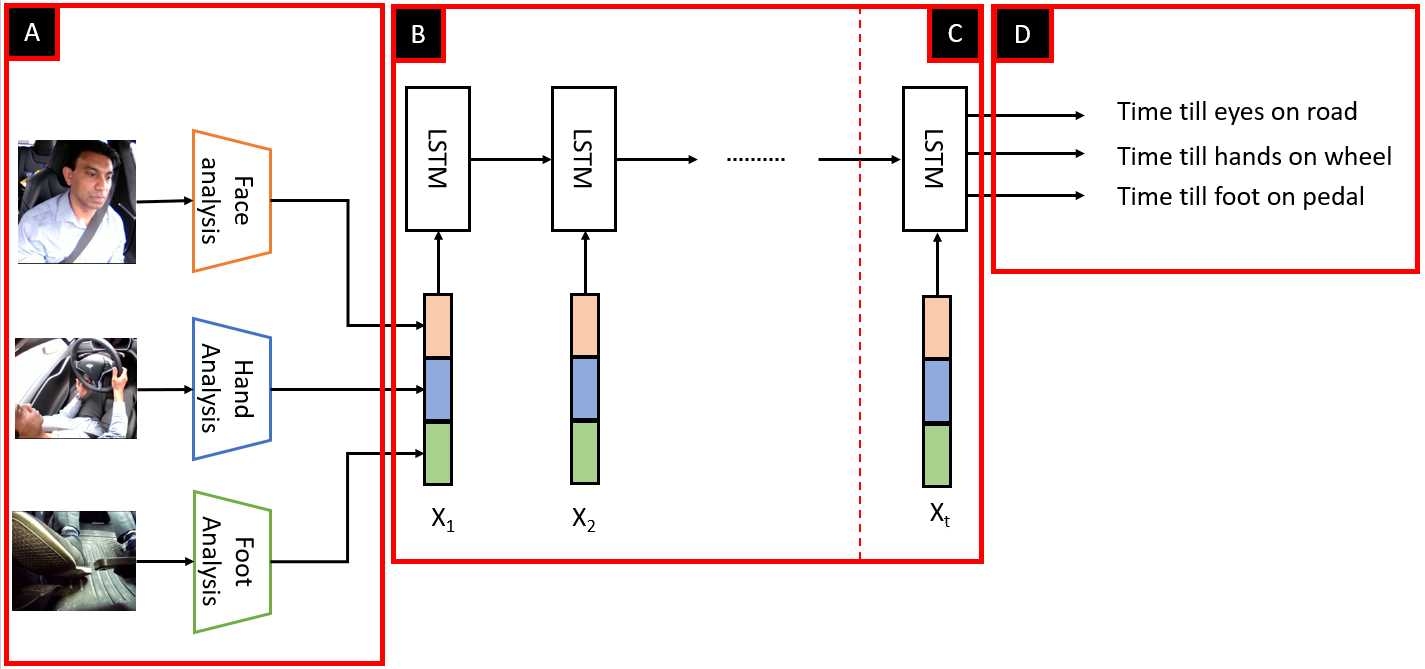}
  \caption{Algorithm workflow for the overall \textit{Take-over Time} prediction framework. Each of the 4 blocks (A-D) above are described in the text.}
  \label{fig:d5-framework_tot}
\end{figure}

\noindent\underline{\textbf{Block A:}} This block comprises of the models operating on \textit{low-level} raw sensory data. These models analyze specific driver behaviors related to the face~\cite{vora2017generalizing, Vora_2018, rangesh2020gpcyclegan}, hands~\cite{yuen2018looking} and the foot~\cite{rangesh2019forced}. Each of these models run as independent processes, on separate threads of execution. These algorithms operate on their respective camera feeds and produce sensor-agnostic mid to high-level features like gaze direction, hand locations, handheld objects, foot locations etc. on a per-frame basis.

\noindent\underline{\textbf{Block B:}} This block comprises of a sequential model responsible for learning features useful for TOT prediction. This downstream model is a recurrent LSTM network, capable of learning both short and long term temporal patterns. This model is run independently, grabbing the latest available mid to high-level features from Block A, processing them, and finally updating its hidden state.

\noindent\underline{\textbf{Block C:}} This block is usually considered part of the previous block, but is highlighted separately for clarity. Block C depicts the latest operation performed by LSTM model described in Block B, and produces the set of outputs for the current (latest) timestep.

\noindent\underline{\textbf{Block D:}} This block lists the post-processed outputs produced by the LSTM model. These outputs are produced at each timestep, not just at the end of a sequence. This makes it possible to get continuous predictions of take-over times for every frame.

We detail the take-over time model introduced in blocks B, C, and D in the following subsection.

\subsection{Take-over Time Model Architecture}
\label{sec:d8-tot-models}








\begin{figure}[h]
\centering
\includegraphics[width=0.8\linewidth]{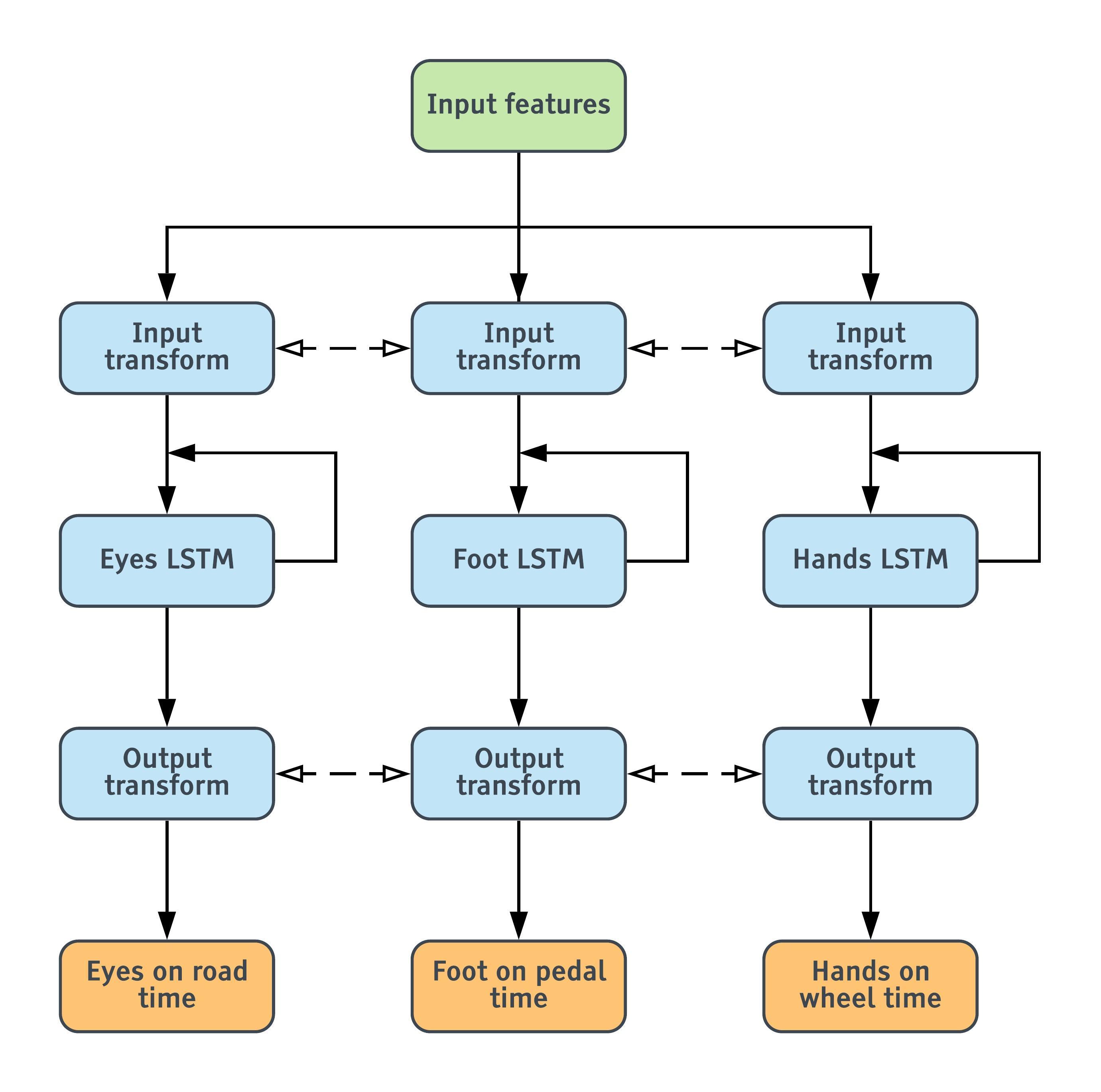}
\caption{The proposed independent LSTMs model architecture.}
\label{fig:d8-id-lstms}
\end{figure}

Instead of directly feeding the input features produced by block A to a recurrent network, we propose to parallely process the input features using separate LSTMs to predict the three times of interest. The reasoning behind this is to accommodate different hidden state update rates for different driver behaviors, for example – eyes-on-road behavior is generally faster (short term) than hands-on-wheel behavior (medium/long term). Having multiple independent LSTMs allows each one to update at different rates, thereby capturing short/medium/long term behaviours separately.

As depicted in Fig.~\ref{fig:d8-id-lstms}, the input features are first transformed using a fully-connected (FC) layer (plus non-linearity), which is then fed to an LSTM at each timestep.
The LSTM layer receives the transformed input features at each timestep and updates its hidden state. In all our experiments, we choose a 2 second window of features as input to our models.
After 2 seconds worth of inputs and updates, the hidden state of the LSTM after the latest timestep is passed through an output transformation (FC layer plus non-linearity) to predict the three times of interest.
Although each branch has its own LSTM cell, the input and output transformations are still shared between the three LSTMs as the feature inputs to the three branches are the same. This tends to reduce overfitting based on our experiments.

We apply a simple $L1$ loss to train this network. Let $o_e$, $o_f$, and $o_h$ be the outputs produced by the model. Assuming $t_e$, $t_f$, and $t_h$ are the target eyes-on-road time, foot-on-pedal time, and hands-on-wheel time respectively, the total loss is:
\begin{equation}
    \mathcal{L} = \frac{1}{N} \sum_{i=1}^{N} |t_e^i - o_e^i| + \frac{1}{N} \sum_{i=1}^{N} |t_f^i - o_f^i| + \frac{1}{N} \sum_{i=1}^{N} |t_h^i - o_h^i|.
\end{equation}
The entire model is trained using an Adam optimizer with a learning rate of $0.001$ for $10$ epochs.

\section{Experiments \& Evaluation}

\subsection{Ablation Experiments}
\label{sec:d8-ablation}

In this subsection, we go through several ablation experiments conducted on the validation set, to evaluate and contrast the importance of various design choices, feature combinations, and training procedures.

First, we conduct an experiment to assess the effectiveness of our trained model to predict take-over components. We use mean absolute errors (MAEs) for each time of interest and the overall TOT as metrics for comparison. Results from these experiments are presented in Table~\ref{tab:data-comp}.

\begin{table}[t]
\centering
\resizebox{\linewidth}{!}{
\begin{threeparttable}\centering
\caption{Prediction errors for different times of interest on the CDS validation set when trained on the augmented dataset.}
\label{tab:data-comp}
\begin{tabular}{@{}ccccc@{}}
\hline
\begin{tabular}[c]{@{}c@{}}Training\\ dataset (s)\end{tabular} & 
\begin{tabular}[c]{@{}c@{}}Eyes-on-road\\ MAE (s)\end{tabular} & 
\begin{tabular}[c]{@{}c@{}}Foot-on-pedal\\ MAE (s)\end{tabular} & 
\begin{tabular}[c]{@{}c@{}}Hands-on-wheel\\ MAE (s)\end{tabular} & 
\begin{tabular}[c]{@{}c@{}}Take-over time\\ MAE (s)\end{tabular}\\
\hline \hline
\rowcolor[HTML]{EFEFEF}
CDS (A)                            & 0.3266 & 0.4841 & 0.7113 & 0.7912 \\
\bottomrule
\end{tabular}

\end{threeparttable}
}
\end{table}

\begin{table}[t]
\centering
\resizebox{\linewidth}{!}{
\begin{threeparttable}\centering
\caption{Prediction errors for different times of interest on the CDS validation set for a variety of feature combinations.}
\label{tab:feat-comp}
\begin{tabular}{@{}ccccccccc@{}}
\toprule
\multicolumn{5}{c}{Features} & \\ \cmidrule(r){1-5}
F\tnote{1} & 
G\tnote{2} & 
H\tnote{3} & 
S\tnote{4} & 
O\tnote{5} & 
 \multirow{-3}{*}{\begin{tabular}[c]{@{}c@{}}Eyes\\ on road\\ MAE (s)\end{tabular}} & \multirow{-3}{*}{\begin{tabular}[c]{@{}c@{}}Foot\\ on pedal\\ MAE (s)\end{tabular}} & \multirow{-3}{*}{\begin{tabular}[c]{@{}c@{}}Hands\\ on wheel\\ MAE (s)\end{tabular}} & \multirow{-3}{*}{\begin{tabular}[c]{@{}c@{}}Take-over\\ time\\ MAE (s)\end{tabular}}\\
\hline \hline
\rowcolor[HTML]{EFEFEF}
 \ding{51} & & & & & 0.3587 & 0.5018 & 0.8599 & 0.8856 \\
 & \ding{51} & & & & 0.3332 & 0.5690 & 0.8411 & 0.8837 \\
\rowcolor[HTML]{EFEFEF}
 & & \ding{51} & & & 0.3729 & 0.5384 & 0.7565 & 0.9012 \\
 & & \ding{51} & \ding{51} & & 0.3783 & 0.5109 & 0.7369 & 0.8315 \\
\rowcolor[HTML]{EFEFEF}
 & & \ding{51} & & \ding{51} & 0.3702 & 0.4973 & 0.7177 & 0.8621 \\
 & & \ding{51} & \ding{51} & \ding{51} & 0.3747 & 0.4857 & 0.7141 & 0.7983 \\
\rowcolor[HTML]{EFEFEF}
 & \ding{51} & \ding{51} & & \ding{51} & 0.3244 & 0.5220 & 0.7163 & 0.7920 \\
 & \ding{51} & \ding{51} & \ding{51} & \ding{51} & 0.3299 & 0.5124 & 0.7215 & 0.7921 \\
 \rowcolor[HTML]{EFEFEF}
\ding{51} & \ding{51} & \ding{51} & \ding{51} & & \textbf{0.3222} & 0.5059 & 0.7870 & 0.8475 \\
\ding{51} & \ding{51} & \ding{51} & & \ding{51} & 0.3277 & 0.5074 & 0.7144 & 0.7918 \\
\rowcolor[HTML]{EFEFEF}
\ding{51} & \ding{51} & \ding{51} & \ding{51} & \ding{51} & 0.3266 & \textbf{0.4841} & \textbf{0.7113} & \textbf{0.7912} \\
\bottomrule
\end{tabular}
\begin{tablenotes}
    \item[1] \textbf{foot features:} probabilities for all 5 foot activities, namely - away from pedal, on break, on gas, hovering over break, and hovering over gas
    \newline
    \item[2] \textbf{gaze features:} probabilities for all 8 gaze zones, namely - front, speedometer, rearview, left mirror, right mirror, over the shoulder, infotainment, and eyes closed/looking down
    \newline
    \item[3] \textbf{hand features:} probabilities for all 6 hand activities (left and right hand), namely - on lap, in air, hovering over steering wheel, on steering wheel, cupholder, and infotainment
    \newline
    \item[4] \textbf{stereo hand features:} distance of left and right hand from the steering wheel
    \newline
    \item[5] \textbf{hand-object features:} probabilities for all 7 hand object categories (left and right hand), namely - no object, cellphone, tablet/iPad, food, beverage, reading, and others
 \end{tablenotes}
\end{threeparttable}
}
\end{table}

Next, we conduct an experiment to assess the relative importance of different input features and their combinations. To isolate effects from features, we train the same independent LSTMs model with different input feature combinations. We use individual and TOT mean absolute errors (MAEs) as metrics for comparison. Table~\ref{tab:feat-comp} contains results from this experiment.

We notice that hand features are the most important, followed by foot and gaze features respectively. This might be because gaze dynamics are relatively predictable during take-overs as the first thing drivers tend to do is look at the road to assess the situation, leading to less variance in eyes-on-road behavior. Next, we notice that adding more informative hand feature like 3D distances to the steering wheel and hand-object information improves the performance further. Hand-objects in particular seem to vastly improve the performance in general. This makes sense as hand-objects provide the strongest cue regarding the secondary activities of drivers. Adding stereo (3D) hand features improves the results, but not by much. Adding foot features also tends to reduce the errors considerably, illustrating the importance of having a foot-facing camera.

In conclusion, one could get close to peak performance by utilizing 3 cameras - 1 foot, 1 hand, and 1 face camera respectively. Hand features (including hand-object features) are most informative, followed by foot and  gaze features respectively.

\subsection{Results \& Analysis}
\label{sec:d8-results}
In this subsection, we provide qualitative and quantitative results of our proposed TOT prediction model to enhance our understanding of their workings and to gain more insights. All results in this subsection are reported on a test set separate from our training and validation splits.

\noindent\underline{\textbf{Quantitative results:}}
First, we present quantitative error metrics on the test set for the proposed model in Table~\ref{tab:test-mae}. To measure the benefits of the independent LSTMs approach, we compare our model to a baseline model with a single LSTM. We observe that the independent LSTMs model outperforms the baseline model across most metrics, and falls slightly short of the baseline for eyes-on-road MAE. We also notice that hands-on-wheel MAEs are usually the largest owing to large variance in hand behaviors and large absolute values associated with hands-on-wheel times.

We also show results for ID LSTMs when trained on 75\% and 90\% of available training data. This helps us gauge the expected improvement in performance as more training data is added. Based on the numbers presented in Table~\ref{tab:test-mae}, we can expect meager improvements as more data is added. This indicates a case of diminishing returns.

\begin{table}[t]
\centering
\resizebox{\linewidth}{!}{
\begin{threeparttable}\centering
\caption{Prediction errors for different models on the take-over time test set.}
\label{tab:test-mae}
\begin{tabular}{@{}ccccc@{}}
\hline
\begin{tabular}[c]{@{}c@{}}Model\\ type (s)\end{tabular} & 
\begin{tabular}[c]{@{}c@{}}Eyes-on-road\\ MAE (s)\end{tabular} & 
\begin{tabular}[c]{@{}c@{}}Foot-on-pedal\\ MAE (s)\end{tabular} & 
\begin{tabular}[c]{@{}c@{}}Hands-on-wheel\\ MAE (s)\end{tabular} & 
\begin{tabular}[c]{@{}c@{}}Take-over time\\ MAE (s)\end{tabular}\\
\hline \hline
\rowcolor[HTML]{EFEFEF}
LSTM\tnote{1}              & \textbf{0.2365} & 0.5007 & 0.8710 & 0.9457 \\
ID LSTMs\tnote{2}          & 0.2497 & \textbf{0.4650} & \textbf{0.8055} & \textbf{0.9144} \\
\hline
\rowcolor[HTML]{EFEFEF}
ID LSTMs (75\%\tnote{3}\ ) & 0.2557 & 0.5013 & 0.8474 & 0.9779 \\
ID LSTMs (90\%\tnote{4}\ ) & 0.2514 & 0.4851 & 0.8482 & 0.9424 \\
\bottomrule
\end{tabular}
\begin{tablenotes}
    \item[1] baseline LSTM model
    \item[2] Independent LSTMs
    \item[3] 75\% of the dataset used for training
    \item[4] 90\% of the dataset used for training
 \end{tablenotes}
\end{threeparttable}
}
\end{table}

\begin{figure}[t]
\centering
\includegraphics[width=0.99\linewidth]{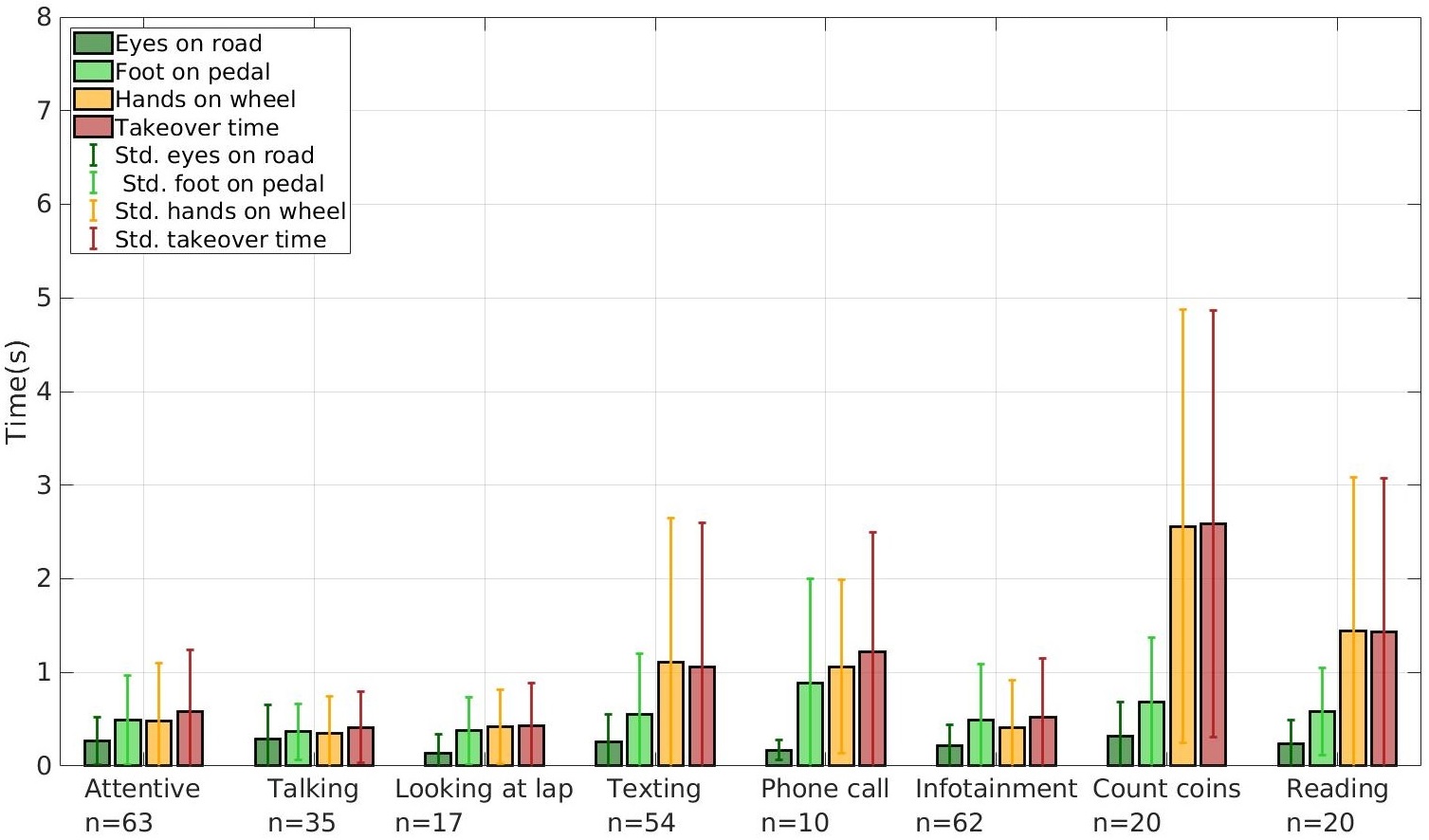}
\caption{Mean absolute error (MAE) per secondary activity in the CDS dataset. The error bars depict one standard deviation in each direction.}
\label{fig:mae-by-sec}
\end{figure}

\noindent\underline{\textbf{Prediction errors by secondary activity:}}
Since our CDS dataset is comprised of a variety of distracting secondary activities, it is useful to conduct error analysis for each activity separately. 
Fig.~\ref{fig:mae-by-sec} shows the MAE values for each secondary activity in the CDS test set. We note that the MAE values have a similar trend as the take-over time statistics (Fig.~\ref{fig:d7.1-totstats}) i.e., activities with larger TOT values have larger MAEs owing to larger variance. However, the errors are typically much smaller than the average take-over times for each secondary activity - indicating small relative errors across all activities. 

\begin{figure*}[t]
\centering
\includegraphics[width=0.99\textwidth]{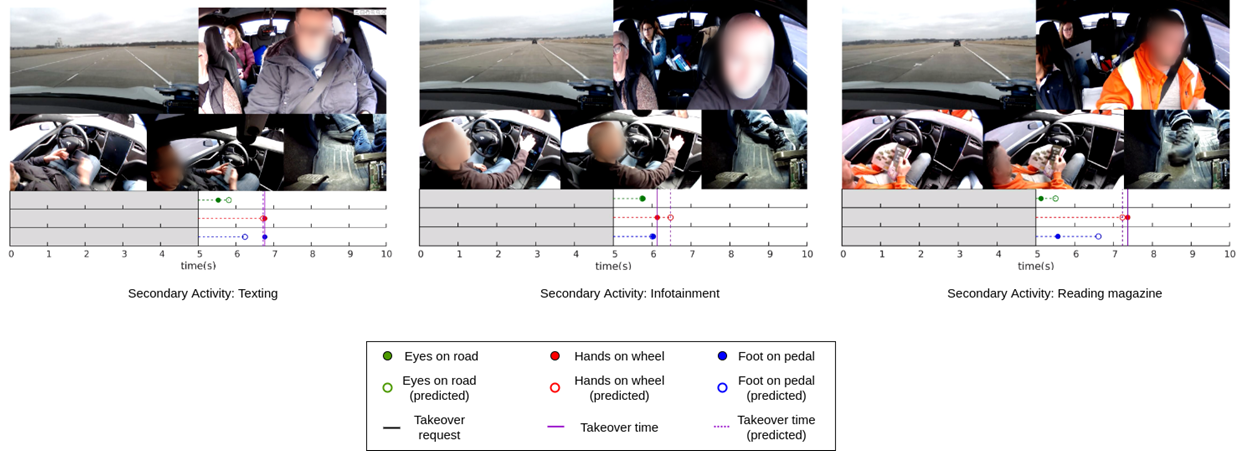}
\caption{Qualitative examples showing predicted take-over times for different secondary activities in the controlled data study (CDS).}
\label{fig:qual}
\end{figure*}

\noindent\underline{\textbf{Qualitative results and examples:}}
Finally, we also provide qualitative examples of predictions made by the ID-LSTMs model for 3 different secondary activities (Fig.~\ref{fig:qual}). Each example shows the 5 camera views at the instant where the TOR is issued. The true values of the 3 times (eyes-on-road, hands-on-wheel, foot-on-pedal) are shown in the plot as solid circular markers, while the corresponding predicted values are shown as hollow circular markers of the same color. We show the the ground truth take-over time as a solid purple line and the predicted take-over time as a dashed purple line.
We note that the model accurately predicts short take-over times when the driver is attentive or operating the infotainment unit, and longer take-over times for activities that impose a higher cognitive load such as texting or reading a magazine. 

\section{Concluding Remarks}
This paper presents one of the largest real-world studies on take-over time (TOT) prediction and control transitions in general. To predict TOTs in the real world, we conduct a controlled data study (CDS) with 89 subjects, each performing multiple take-overs while engaged in a variety of distracting secondary activities. The CDS was used to capture real-world take-over events and corresponding take-over times for a diverse pool of drivers.
This dataset of take-over times was then used to train downstream models for TOT prediction that operate sequentially on mid to high-level features produced by computer vision algorithms operating on different driver-facing camera views. 
In addition to the TOT prediction model, we also provide results from various ablation studies to compare and contrast the effects of different model architectures, feature combinations, and other design choices. We believe that this study outlines the sensors, datasets, algorithms and models that can truly benefit the intermediate levels of automation by accurately assessing driver behavior and predicting take-over times - both of which can then be used to smoothly transfer control between humans and automation.

\section{Acknowledgments}
We would like to thank the Toyota Collaborative Safety Research Center (CSRC) and other sponsors for their generous and continued support. We would also like to thank our colleagues at the Laboratory for Intelligent and Safe Automobiles (LISA), UC San Diego for their useful inputs and help in collecting and labeling the dataset. 

\bibliographystyle{IEEEtran}
\bibliography{root}

\end{document}